\documentclass[runningheads]{llncs}
\usepackage{graphicx}
\usepackage{amsmath,amssymb} 
\usepackage{color}
\usepackage{multirow}

\begin{document}

\title{Towards Locally Consistent Object Counting with Constrained Multi-stage Convolutional Neural Networks
\thanks{\textbf{Acknowledgments} This work was partly funded by NSFC(No.61571297, No.61420106008), the National Key Research and Development Program (2017YFB1002401), and STCSM(18DZ2270700).}} 
\titlerunning{Object Counting with Constrained Multi-stage CNN} 


\author{Muming Zhao\inst{1,2} \and
Jian Zhang\inst{2} \and
Chongyang Zhang\inst{1} \and
Wenjun Zhang\inst{1} }
%

\authorrunning{M.Zhao et al.} 


\institute{Shanghai Jiao Tong University, Shanghai 200240, China \\
\email{sunny\_zhang@sjtu.edu.cn} \\ \and
University of Technology, Sydney, NSW 2007, Australia 
\\ 
}

\maketitle

\begin{abstract}
High-density object counting in surveillance scenes is challenging mainly due to the drastic variation of object scales. The prevalence of deep learning has largely boosted the object counting accuracy on several benchmark datasets. However, does the global counts really count? Armed with this question we dive into the predicted density map whose summation over the whole regions reports the global counts for more in-depth analysis. We observe that the object density map generated by most existing methods usually lacks of local consistency, i.e., counting errors in local regions exist unexpectedly even though the global count seems to well match with the ground-truth. Towards this problem, in this paper we propose a constrained multi-stage Convolutional Neural Networks (CNNs) to jointly pursue locally consistent density map from two aspects. Different from most existing methods that mainly rely on the multi-column architectures of plain CNNs, we exploit a stacking formulation of plain CNNs. Benefited from the internal multi-stage learning process, the feature map could be repeatedly refined, allowing the density map to approach the ground-truth density distribution. For further refinement of the density map, we also propose a grid loss function. With finer local-region-based supervisions, the underlying model is constrained to generate locally consistent density values to minimize the training errors considering both the global and local counts accuracy. Experiments on two widely-tested object counting benchmarks with overall significant results compared with state-of-the-art methods demonstrate the effectiveness of our approach. 

\end{abstract}
\section{Introduction}
\label{chaper_introduction}
Automatic object counting in images using computer vision techniques plays an essential role in various real-world applications such as crowd analysis, traffic control and medical microscopy~\cite{sindagi2017survey}, and hence has gained increased attention in recent years. Currently, the density-map-estimation based counting framework~\cite{lempitsky2010learning} learns to regress a spatial object density map instead of directly estimating the global counts, which is further reported by the summation of pixel values over the whole region on the density map. Due to its effective exploitation of the spatial information, this paradigm has been adopted by most later counting methods~\cite{fiaschi2012learning,pham2015count}. Recently the prevalence of deep learning combined with the density-map-estimation paradigm has largely boosted the counting accuracy on several benchmark datasets~\cite{zhang2015cross,onoro2016towards,zhang2016single,sam2017switching,sindagi2017generating,xie2018microscopy}. However, does the global counts really count? Our observation is that despite the improved global counting accuracy, significant local counting errors exist when diving into the predicted density map. This phenomenon has been reported in~\cite{guerrero2015extremely,sindagi2017survey}, however, it has not been sufficiently investigated and explicitly addressed. 
\begin{figure}[t]
\centering
\includegraphics[width = 1\columnwidth]{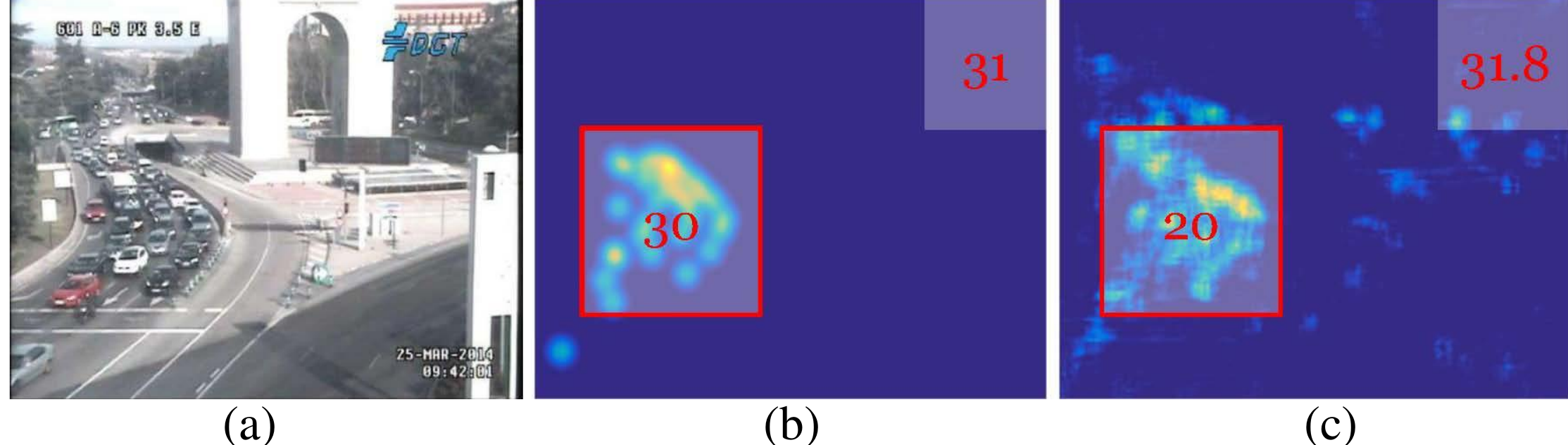}
\caption{Illustration of a locally inconsistent density map prediction. (a) to (c): the original image, the ground truth and the estimated density map. We observe that although the estimated total count (shown in the upper right box) is very close to the ground truth, the quality of prediction is not satisfactory with observation of obvious background noise and count errors of local regions (shown in the red-line-framed boxes).}
\label{fig_intro}
\end{figure}

Here we term this problem as \textbf{\emph{local inconsistency}}. This is to denote the fact that, although a predicted density map can report accurate global count for an input image, the quality of prediction is not good from local perspectives: errors arise when counting objects in subregions of the image. This can be mainly attributed to the various object scales for most images taken in surveillance scenes with perspective distortion. With this property, the model is usually difficult to generate density values which adapt to the drastic changing scales. An example of a locally inconsistent prediction of density map is shown in Fig.~\ref{fig_intro}. It can be observed that the estimated global count (\emph{31.8}) is very close to the ground-truth(\emph{30}). However, errors are exposed to the selected ROI and background regions. For the ROI area with objects, the predicted local count  is only \emph{20}, which is far more satisfactory compared to its real value (30). At the same time, the predicted count (\emph{11.8}) for the background region takes a nearly 30\% proportion of the estimated global count (\emph{31.8}), whose influence to the counting accuracy should not be neglected. The Existence of local inconsistency of the predicted density map not only degrades the reliability of the finally reported object count, and also limits the quality of predicted object density distribution for related higher-level tasks~\cite{sindagi2017survey}. In Section~\ref{proof} we mathematically demonstrate that for an image the local object counting errors decide the upper bound of the global counting errors. In this way, pursing a locally consistent density map which aims to decrease local counting errors as much as possible is a reliable way to help improve the global counting accuracy. 

In this paper, we start from this observation of locally inconsistent problem and propose a joint solution from two aspects. Current existing CNN-based methods handle object scale variations mainly by engineering multi-scale features either with multi-column architectures~\cite{zhang2016single,sam2017switching,Kumagai2018} or with multi-resolution inputs~\cite{onoro2016towards}. We differently exploit a simple yet effective stacking formulation of plain CNNs. Benefited from the internal multi-stage learning process, the feature map is repeatedly refined, and the density map is allowed to correct its errors to approach the ground-truth density distribution. The multi-stage network is fully convolutional and can generate corresponding-sized density map for an arbitrary-sized input image. We also propose a grid loss function to further refine the density map. With finer local-region-based supervisions, the model is constrained to generate locally consistent density values to help minimize the global training errors. The grid loss is differentiable and can be easily optimized with the Stochastic Gradient Descent (SGD) algorithm. 

We summarize our main contribution as follows:
\begin{itemize}
\item For the observed local inconsistency problem, we propose a constrained multi-stage Convolutional Neural Networks (CMS-CNN) for jointly handling from two aspects. 
\item We exploit the multi-stage formulation to pursue locally consistent density map through repeatedly evaluation and refinement, and we also propose a grid loss function to further constrain the model to satisfy the demanding of locally consistent density values.
\item Experiment results on two widely-adopted datasets demonstrated the effectiveness of the proposed method. 
\end{itemize} 

\section{Related Work}
Traditional detection-based counting methods mainly rely on the performance of object detection algorithms~\cite{sidla2006pedestrian,lin2010shape,gao2016people} and are usually fragile especially in crowded scenes with limited object sizes and severe occlusions. Alternatively, early regression approaches directly learn a mapping function from foreground representations to the corresponding counts~\cite{kong2006viewpoint,chen2012feature}, avoiding explicit delineation of individuals. However, this global regression approach ignores the useful spatial information. Towards this goal, a novel framework is proposed in the seminal work~\cite{lempitsky2010learning}, which formulates object counting as a spatial density map prediction problem. With a continuously-valued density assigned for every single pixel, the final object counts can be obtained by summation of the pixel values over the whole density map. Enabling the utilization of spatial information, counting by density map prediction has been a widely-adopted paradigm for later counting approaches~\cite{fiaschi2012learning,pham2015count}. However, the representation ability of hand-crafted features limits the performance of those methods, especially for more challenging situations with severe occlusions and drastic object scale variations. Following our analysis in Section~\ref{chaper_introduction}, our solution to the local inconsistency problem is related to those counting methods handling the object scale variation, which we will give a detailed discussion below.

Recently the prevalence of deep learning technique has largely boosted the counting performance~\cite{zhang2015cross,zhang2016single,onoro2016towards,sindagi2017generating,xie2018microscopy,sam2017switching}. Along with several newly-emerged datasets~\cite{guerrero2015extremely,zhang2015cross,zhang2016single,chan2008privacy} which contain extremely dense crowd in clutter background with large perspective distortion, the drastic object scale variation has been one of the most important problems which hinders the counting accuracy. Several deep learning based counting methods are proposed to deal with this situation. To overcome the object scale variations, one of the earliest works~\cite{zhang2015cross} propose a patch-based convolutional neural network (CNN) to process normalized patches for object density map estimation. Given an image, patches are extracted at a size proportional to their perspective values and then are normalized into one same scale. In this way, object scale distinctions in the input patches are alleviated during the training data preparation stage, which relieves the burden applied to the regression model on various scale handling. However, with the normalization procedure objects in the image may be distorted~\cite{he2014spatial}. To improve training and inference efficiency, later methods shift their focus to in-network handling of the scale variation. In~\cite{zhang2016single} a multi-column neural network~\cite{cirecsan2012multi} (MCNN) is proposed to incorporate multi-scale features for density map estimation. The network consists of three columns with small, medium and large kernel sizes respectively. Feature maps from the three sub-models are then aggregated to generate the final object density map. To further improve the multi-scale feature fusion efficiency, an improved work is proposed in~\cite{Kumagai2018} where the MCNN is viewed as three experts to handle objects in each of the three scales respectively, and another gating CNN is proposed to decide the confidence of the generated feature map by each expert for an input image. The confidence is learned and formulated by the gating CNN as the weights applied for feature map fusion to estimate final object counts. Furthermore, in~\cite{sam2017switching} the authors propose to leverage the internal object density distinctions and assign the three columns of networks in MCNN to process image patch other than the whole image. A switch network is proposed to relay each image patch to the best-suited model for density map estimation. With a similar strategy to the MCNN~\cite{zhang2016single}, in~\cite{onoro2016towards} a Siamese CNN is employed to receive multi-resolution input to generate corresponding multi-scale features for density map estimation. 

To the best of our knowledge, most existing methods on object scale variations handling mainly rely on the formulation of multi-scale feature either with the multi-column architecture~\cite{cirecsan2012multi} or using the multi-resolution input~\cite{onoro2016towards}. In this paper, we start from our observation with the local inconsistency problem and propose a joint solution from two aspects. First, we resort to a completely different formulation with previous methods which stacks multiple plain CNNs to handle the scale variation. Benefited from the internal multi-stage inference mechanism~\cite{newell2016stacked}, the feature is repeatedly evaluated for refinement and correction, allowing the estimated density map to approach the ground-truth density distribution gradually with local consistent density values. In the other aspects, we propose a grid loss function to further constrain the model to adjust density values that are not consistent with local object counts. The multi-stage mechanism has been proven effective in various computer vision tasks like face detection~\cite{qin2016joint}, semantic segmentation~\cite{li2016iterative}, and pose estimation~\cite{newell2016stacked}. In this paper, we exploit the multi-stage mechanism with the proposed grid loss towards locally consistent object counting. Our model is trained end-to-end efficiently, with validated effectiveness on two publicly available object counting datasets. 

\section{Relationship Between Global Counting Errors and Local Counting Errors}
\label{proof}
Given a pair of ground truth and predicted density map $\left \{D_{gt}, D_{es} \right \}$ of an image $I$, we manually divide the map into $T$ non-overlap grids denoted as $B = \left\{b_{1}, b_{2},\cdots, b_{T}\right \}$. Mean Absolute Error (MAE) is used to measure the global counting accuracy, i.e., $E_{I} = \left|\sum_{i=1}^{N}D_{gt}(p_{i}) - \sum_{i}D_{es}(p_{i})\right|$, where $N$ is the pixel number in image $I$. Reformulate above equation in terms of subregions will obtain:
\begin{align}
E_{I} &= \left|\sum_{j=1}^{T}\sum_{i\in b_{j}}(D_{gt}(p_{i}) - D_{es}(p_{i}))\right| \\
&= \left|\sum_{j=1}^{T}E_{b_{j}}\right| \leq \sum_{j=1}^{T}\left| E_{b_{j}}\right|,
\label{eq_relation_global_local}
\end{align}
\noindent where $E_{b_{j}}$ denotes the MAE of the object count in local region $b_{j}$. From Eqn.~\eqref{eq_relation_global_local} it can be concluded that summation of MAE of object counts in each non-overlap subregions is an \emph{upper bound} of the MAE of the global object counts in the whole image. From this perspective, pursuing a locally consistent density map which aims to decrease local counting errors will help improve the reliability as well as drive the accuracy of the global object counts.

\section{Constrained Multi-stage Convolutional Neural Networks}
Our overall model consists of two components, \emph{multi-stage convolutional neural network} and the \emph{grid loss}. Since the grid loss provides additional supervisions, it can be viewed as constraints to the proposed multi-stage network. Before presenting the details, we first give the formulation of density-map-prediction based object counting paradigm. 
\subsection{Density Map Based Object Counting}
In this work, we formulate the object counting as a density map prediction problem~\cite{lempitsky2010learning}. Given an image $I$ with the dotted annotation set $A_{I}$ for target objects, the ground truth density map $D_{gt}$ is defined as the summation of a set of 2D Gaussian functions centered at each dot annotation, i.e., $\forall p \in I, D_{gt}(p) = \sum_{\mu\in A_{I} }\mathbb{N}(p;\mu,\Sigma)$, where $\mathbb{N}(p;\mu,\Sigma)$ denotes a normalized 2D Gaussian kernel evaluated at $p$, with mean $\mu$ on each object location and isotropic covariance matrix $\Sigma$. Total object count $C_{I}$ for image $I$ can be obtained by summation of pixels' values over the density map. Note that all the Gaussian are summed to preserve the total object count even when there are overlaps between objects~\cite{onoro2016towards}.
\begin{figure*}[t]
\centering
\includegraphics[width = 0.9\textwidth]{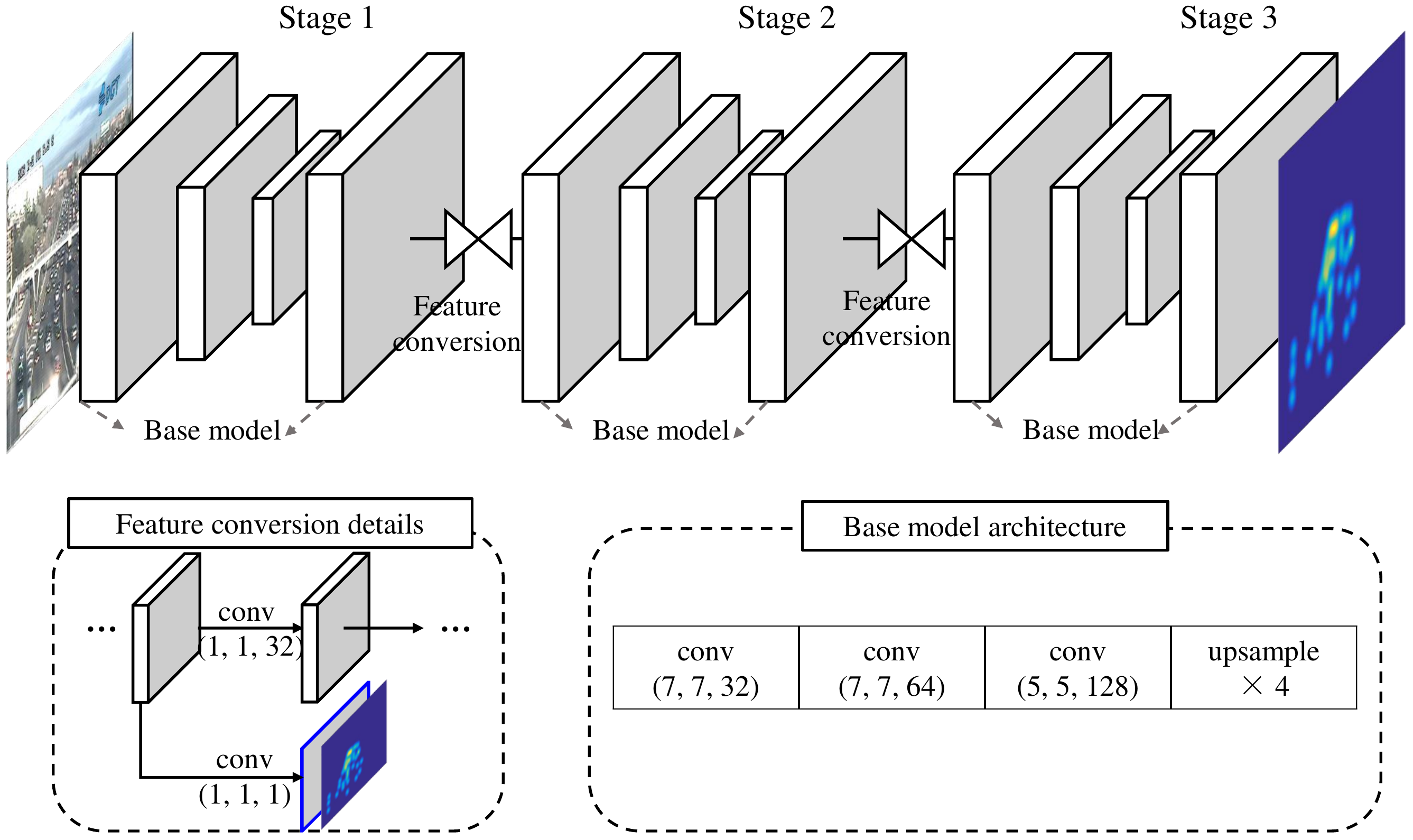}
\caption{Architecture of the multi-stage convolutional neural network. We stack several base models sequentially with feature conversion blocks which i). perform feature dimension alignment of feature maps between two adjacent base models, and ii). generate a prediction
for each base model to enable intermediate supervision. The first base model accepts the input image, and the rest base models in the following stages accept feature maps which comes from the previous feature conversion block.
}
\label{fig_method}
\end{figure*}

Given this counting framework, the goal of our work is to learn a mapping function from an input image $I$ to its estimated object density map $D_{es}$, i.e., $\forall p \in I, D_{es}(p) = F(p|\Theta)$, where the underlying model is parameterized by $\Theta$. 
\subsection{Multi-stage Convolutional Neural Network}
To generate locally consistent density values, we resort to the stacking formulation of plain CNNs. We exploit the internal multi-stage inference mechanism to repeatedly evaluate the feature map and allow the generated density map to be refined to figure out the best-suited density values. Mathematically, 
For each pixel $p$ in the training image $I$, we learn the mapping function $F(p|\Theta)$ in a compounded way with a series of functions from different stages:
\begin{equation} 
\label{eq:compound}
F(p|\Theta) = f_K(\cdot|{W}^K) \circ \cdots \circ f_s(\cdot|{W}^s) \circ \cdots \circ f_2(\cdot|{W}^2) \circ f_1(\cdot|{W}^1),
\end{equation}
\noindent where \{$f_{s}, s = K, K-1, \cdots 2, 1$\} represents the base model parameterized by $W^{s}$ in the $s$ stage, and $\circ$ denotes the function compounding operation. With this decomposition, we can add intermediate supervisions~\cite{lee2015deeply} to each base model $f_s$ to facilitate the training process. A pixel-wise $L2$-norm loss function can be applied for training: 
\begin{align}\label{pixel-level loss}
\begin{split}
L(W, D_{es}) & = \frac{1}{N} \sum_{p} \sum_s \alpha_s \left \| D_{es}^{s}(p)-D_{gt}(p) \right \|_{2}^{2}, \\
&  = \frac{1}{N} \sum_{p} \sum_s \alpha_s L^s_p   
\end{split}
\end{align}
\noindent where $D_{es}^{s} = f_{s}(X^{s-1}|\widehat{W}^{s})$ is a side output density map of base model $f_s$, $X^{s-1}$ are feature maps produced by the model $f_{s-1}$ in the previous stage, $W = \{W^{s}, \widehat{W}^{s}\}_{s = 1,\cdots, K}$ are parameters of the whole model, $N$ is the number of pixels in image $I$ and $\alpha_s$ is the weight for the side output loss of base model $f_s$. 

Fig.~\ref{fig_method} illustrates the proposed multi-stage model, where the base model is formulated as a fully convolutional neural network~\cite{long2015fully}. For a convolution (conv) layer, we use the notation ($h$, $w$, $d$) to denote the filter size $h\times w$ and the number of filters $d$. Inspired by~\cite{zhang2015cross} the convolution part of our base model contains three convolution layers with sizes of (\emph{7, 7, 32}), (\emph{7, 7, 64}) and (\emph{5, 5, 128}) respectively, each followed by a ReLu layer. Max Pooling layer with $2\times 2$ kernel size is appended after the first two convolution layers. Considering the input image is downsampled by a stride of 4, we add a deconvolution layer at the end of each base model to perform in-network upsampling to recover the original resolution. The resulted feature maps of each base model are fed into the subsequent stage after dimension alignment with a $1\times 1$ convolution layer of the feature conversion block. Inspired by the success of training CNN models with deep supervisions~\cite{lee2015deeply,newell2016stacked}, another $1\times 1$ convolution layer is appended on the feature maps to predict a side output of density map, where the intermediate supervision will be then applied. Applying supervisions on each base model help facilitate the learning process of the whole network. The feature conversion and intermediate supervision block are illustrated in Fig.~\ref{fig_method}. Except for the first base model that accepts the input image, the first convolution layers of the following base models are modified to be consistent with the dimensions of previously generated feature maps. 

\subsection{Grid Loss}
To further refine the density map to generate accurate global counts as well as the local counts, we also propose a grid loss function as the supervision signal. With the consideration of training error in local regions, the model is constrained by the grid loss to correct those density values which result to severely conflicts of estimated local counts with the ground truth. 

\begin{figure}[t]
\centering
\includegraphics[width = 1\columnwidth]{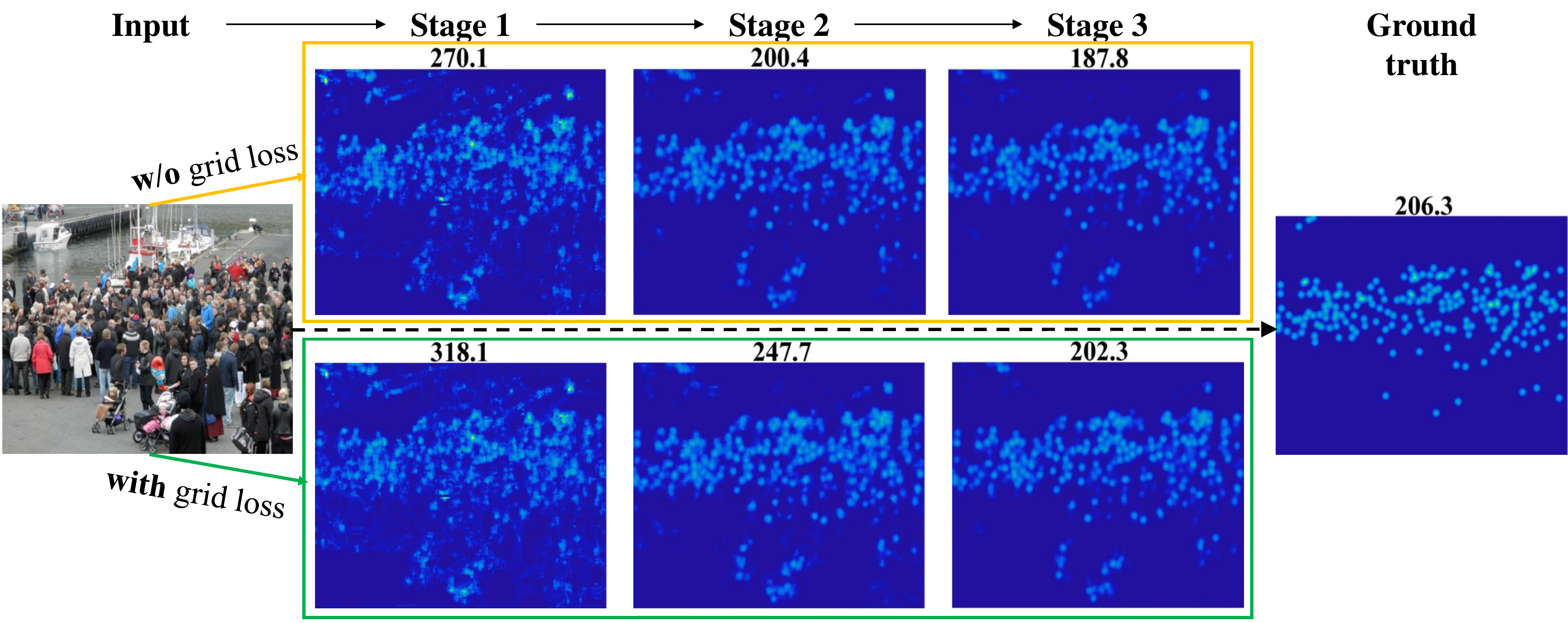}
\caption{Effects of the grid loss on a three-stage model. It can be observed that training with grid loss drives the model to learn to correct the regression errors and produce more accurate object counting results.} 
\label{fig_grid_loss_effects}
\end{figure}

Divide an image into several non-overlapping grids, and the grid loss can be depicted with local counting errors in each sub-region. The traditional pixel-wise loss (Eqn.~\eqref{pixel-level loss}) measures pixel-level density divergence while the grid loss reflects region-level counting difference. Considering the numerical gap between the numerical value between the global and local counting errors, we depict the grid counting loss with the average density loss for pixels within each specific area. This is based on the assumption that within a relatively small area, it has a great chance that pixels' density values are very similar. Then it can be regarded that every single pixel within this area has a density loss which contributes to the total count loss. By distributing the total count loss to each pixel, the grid loss help drive the correction of most violated density values and improve regression accuracy. Following previous notation in Eqn.~\eqref{pixel-level loss}, for a group of non-overlap grid set $B = \left \{b_{1},b_{2},\cdots,b_{T}\right\}$ in the predicted density map $D_{es}$, the grid loss is defined as
\begin{equation}
L_{grid} = \sum_{j=1}^{T}\left\|\frac{1}{\left|b_{j}\right|}(\sum_{p\in b_{j}}D_{es}(p)-\sum_{p\in b_{j}}D_{gt}(p)) \right\|_{2}^{2},
\label{grid_loss_1}
\end{equation}

\noindent where $\left|b_{j}\right|$ denotes the pixel number in this grid. Reformulation of the grid loss for the multi-stage model will be
\begin{equation}
\begin{split}
\widehat{L}^s_p = (1-\lambda^s)L^s_p + \lambda^s L^s_{grid},
\end{split}
\label{grid_loss_2}
\end{equation}
\noindent where $\lambda^s$ is a weight scaler applied to trade off between the estimator, i.e., the traditional pixel-wise loss and the modulator, i.e., the proposed grid loss. Substitute $L^{s}_p$ in Eqn.~\eqref{pixel-level loss} with Eqn.~\eqref{grid_loss_2} will derive the final grid loss used to supervise the whole network. With this formulation, it can be observed that each pixel is not only supervised by the original density loss, and is also additionally regularized by the average density loss of the block it belongs to. This will drive the model to correct those density values that are not consistent with local object counts and improve final counting accuracy. 
In Fig.~\ref{fig_grid_loss_effects} a sample image is given to show the effects of the grid loss on a three-stage model. It can be seen that training the multi-stage model with grid loss is able to drive the model to correct regression errors and obtain more accurate object counts. 

\section{Experimental Results}
\subsection{Experiment Setting}
\noindent\textbf{Training Details}
Our model is implemented using MatConvNet~\cite{vedaldi2015matconvnet} with the SGD optimization. The hyper-parameters of our network include the mini-batch size (64), the momentum (0.9) and the weight decay ($5$$\times$$10^{-4}$). We experimentally set the weights for each intermediate supervision $\alpha_{i}$ to be $1$ and the weight for grid loss $\lambda_{i}$ to be $0.5$. Detailed analysis of $\lambda_{i}$ in the grid loss is given in Section~\ref{param selection}. The grid size is set according to the average object size in the dataset, and is $56$ for an $224$$\times$$224$ image.  Training starts from an initial learning rate of $1$$\times$$10^{-6}$, which is divided by 10 after the validation loss plateaus.

\noindent\textbf{Model Initialization}
Considering the difficulty to train a deep model from scratch, we take advantage of the widely-used pre-training strategy. The base CNN model is first trained and then is duplicated to construct the multi-stage network. Additional weights, e.g., the feature alignment layers between adjacent base models are randomly initialized. Finally, the whole model is fine-tuned end-to-end. 

\noindent\textbf{Data Augmentation}
During training, 20 image patches with a size of 224$\times$224 are randomly cropped from each training image for data augmentation. Randomly flipping and color jitter are performed for data augmentation. Note that the ground truth density map is a combination of 2D Gaussian functions, and their numeric values are very small ($10^{-3}$$\sim$$10^{-5}$) to enable effective learning. For this reason, we magnify the ground truth density map by a factor of 100 during the training process. 

\noindent\textbf{Running Time}
With end-to-end training, it takes about 15 hours to train a 3-stage CNNs on a single NVIDIA TITAN X GPU. For testing it takes about 0.15$s$ for an image of size $576$$\times$$720$.

\subsection{Evaluation Metrics}
Given a test image $I$, we directly use the output from the last stage of the network as the density map prediction. Three standard metrics are utilized for evaluation: mean absolute error (MAE), mean square error (MSE), and the grid average mean absolute error (GAME). For a dataset with $M$ test images the MAE is defined as $MAE= \frac{1}{M}\sum_{i=1}^{M}\left|C_{es}^{i}-C_{gt}^{i}\right|$, where $C_{es}^{i}$ and $C_{gt}^{i}$ are the predicted and the ground truth object counts for the $i$-th image. MSE measures the robustness of the predicted count, which is defined as $MSE = \sqrt{\frac{1}{M}\sum_{i=1}^{M}(C_{es}^{i}-C_{gt}^{i})^{2})}.$

MSE and MAE evaluate the global object counts while ignoring the local consistency of predicted density maps. We additionally include the Grid Average Mean Absolute Error (GAME)~\cite{guerrero2015extremely} as a complementary evaluation metric. After dividing a density map into $4^{L}$ non-overlapping regions, GMAE for level $L$ is defined as:
\begin{equation}
GAME(L) = \frac{1}{N}\cdot \sum_{i=1}^{M} \left ( \sum_{l=1}^{4^{L}}\left | C_{es}^{il}-C_{gt}^{il} \right | \right ),
\label{game}
\end{equation}
\noindent where $C_{es}^{il}$ and $C_{gt}^{il}$ denotes the predicted and ground truth counts within the region $l$ respectively. The higher $L$, the more restrictive this GAME metric will be on the local consistency of the density map. Note that the MAE metric is a special case of GAME when $L=0$.
\subsection{Hyper-Parameter Selection in Grid Loss}
\label{param selection}
There are three hyper-parameters in the proposed grid loss function: the grid size, the loss weights $\alpha$ for each base model and the weights $\lambda$ to balance the pixel-wise loss and grid loss. We experimentally fix $\alpha$ to be 1 across different stages and study the effects of another two parameters. The grid size denotes the number of blocks divided in the image. The weighting scaler $\lambda$ is in charge of the modulation degree of a block count loss on its inner pixels. We conduct experiments comparing the MAE of applying grid loss to a 2-staged model with different hyper-parameter settings of grid size $1, 2, 4, 8, 16$ and $\lambda = 0.9, 0.5, 0.1, 0.01$. 

Experimental results show our method performs best with $\lambda = 0.5$ and block size of 4. We use this setting across all our experiments unless otherwise specified. $\lambda = 0.9$ degrades the original performance for almost all the grid size settings, which implies that large weighting scaler may disturb the normal density learning process. As $\lambda$ further decreases, the network converges to the performance training with the pixel-wise loss. When the grid size is too big, each grid area will become too small to effectively include objects, and the performance starts to degrade to the per-pixel density loss. 
\begin{table}[!htbp]
\centering
\caption{Performance of ablation experiments for network structures and supervisions.
}
\label{ablation study}
\begin{tabular}{c|c|c|c}
  \hline
  index & Design choices & MAE & MSE \\
  \hline
  \hline
$a.$ & MS-CNN-1 (the base model) & 107.7  & 173.2  \\
$b.$ & CMS-CNN-1 & 101  & 160.4 \\ 
$c.$ & MS-CNN-2 & 82.3  & 140.4  \\
 $d.$ & CMS-CNN-2 & 74.2  & 127.6  \\
  $e.$ & MS-CNN-3 & 74.4  & 129.7 \\
   $f.$ & CMS-CNN-3 & 73 & 128.5  \\
  \hline
\end{tabular}
\end{table}
\subsection{Ablation Experiments}
We perform extensive ablation experiments on ShanghaiTech Part-A dataset to study the role of the multi-stage convolution network and the grid loss separately play in the whole constrained multi-stage networks. Results of alternative design choices are summarized in Table~\ref{ablation study}. For simplicity, we denote the multi-stage model with $n$ stages as MS-CNN-$n$, and the corresponding constrained model trained with grid loss as CMS-CNN-$n$.

From Table~\ref{ablation study} several observations could be drawn. First, the multi-stage formulation of plain CNNs (compare between $a$, $c$, $e$) and the proposed grid loss (compare $a$ and $b$, $c$ and $d$, $e$ and $f$) both demonstrate effectiveness in improving counting accuracy. Second, the overall MAE performance of the constrained multi-stage CNNs (CMS-CNN) can be improved by adding stage by stage. We observe the MSE performance of MS-CNN-3 degrades the performance of CMS-CNN-2 a little bit. We suspect this may be the reason that with more stages added, the model becomes deeper to be well optimized. 

\subsection{Comparison with the State-Of-The-Arts}
\label{comparison section}
\noindent \textbf{ShanghaiTech} The ShanghaiTech dataset~\cite{zhang2016single} is a large-scale dataset which contains 1198 annotated images. It is divided into two parts: there are 482 images in part-A and 716 images in part-B. Images in part-A are collected from the Internet and the part-B are surveillance scenes from urban streets. We follow the official train/test split~\cite{zhang2016single} which is 300/182 for part-A and 400/316 for part-B. For validation, about 1/6 images are randomly selected from the original training data to supervise the training process. 

\begin{figure}[!htbp]
\centering
\includegraphics[width = 1\columnwidth]{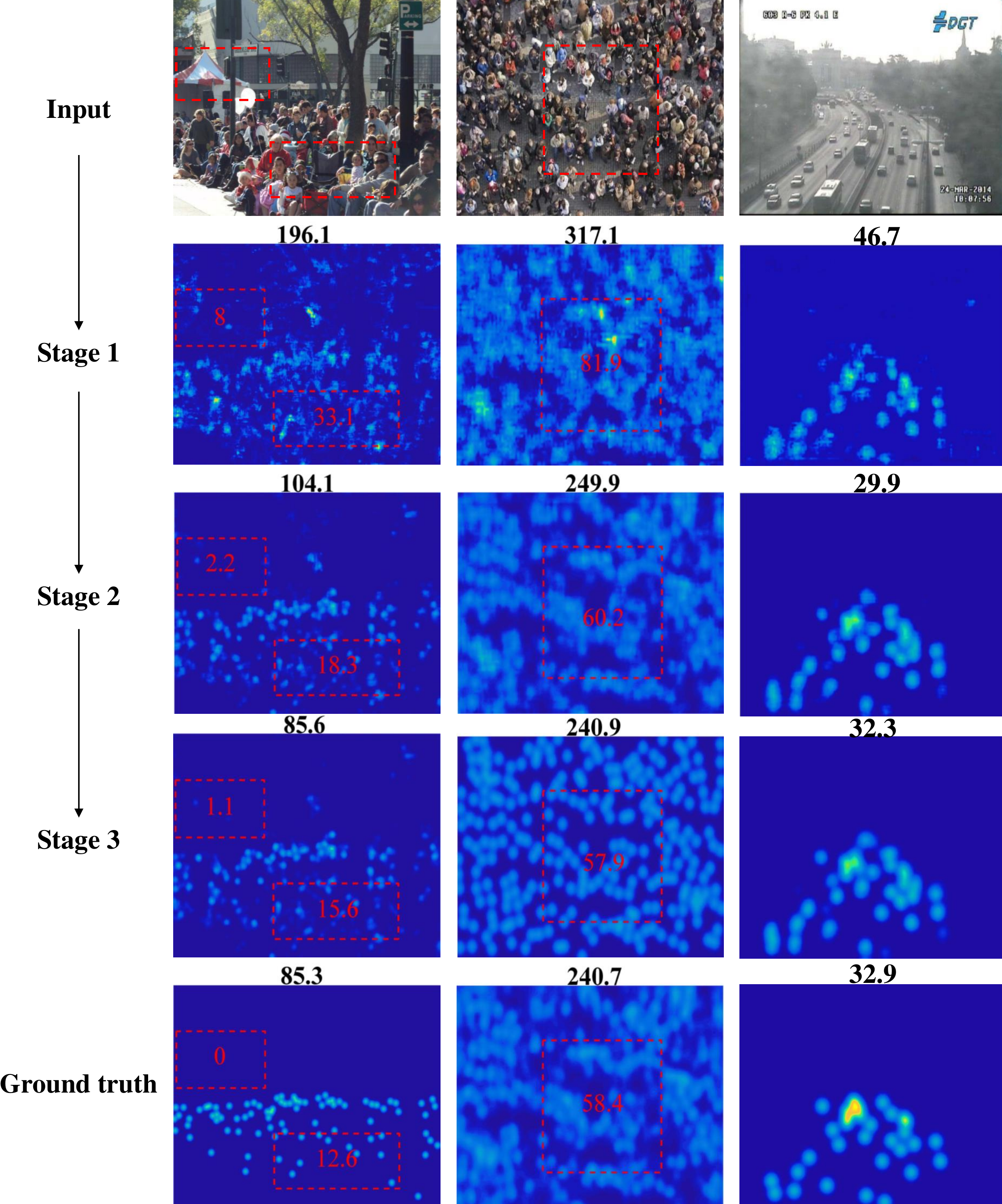}
\caption{Density map prediction results as input images proceed through the multi-stage convolution model. The first row lists images sampled from the ShanghaiTech dataset (first two) and the TranCos dataset (last one). The second to the fourth rows show the intermediate outputs from the first two stages and the final prediction of the last stage, respectively. The ground truth density maps are shown in the last row. Object count derived from the density map are labeled on top of each prediction result. For the first two crowded sample images we also randomly select several subregions to track the local object counts, which are shown in the red boxes.} 
\label{fig_subsequent_results}
\end{figure}

Table~\ref{game_sha} reports the comparison results with five baseline methods: Crowd-CNN~\cite{zhang2015cross}, MCNN~\cite{zhang2016single}, Cascaded-MTL~\cite{sindagi2017cnn}, Switch-CNN~\cite{sam2017switching}, CP-CNN~\cite{sindagi2017generating}. On Part-A our methods achieves best MAE among all the comparison methods, and the second-best MSE. We observed that most images in Part-A are extremely crowded and also have pretty uniform object scales within the image, where the context information matters much compared to considering object scale variations to derive accurate counting results. In~\cite{sindagi2017generating} the counting method is proposed from the perspective of context information modeling, which better suits the situation on Part-A. On Part-B our method outperforms all other methods and evidences a 40\% improvements in MAE over CP-CNN~\cite{sindagi2017generating}. Fig.~\ref{fig_subsequent_results} illustrates the inference process in each stage with the CMS-CNN-3 model of two sample images from ShanghaiTech dataset. For the first image, it can be observed that the total object counts gradually approaches the ground truth. What's more, errors exist in the upper left background region are gradually refined and the local counting accuracy is also gradually improved. The similar situation can be observed for the second image, where the predicted density map is becoming more consistent with the ground-truth density distributions. 
\begin{table}[!thbp]
\centering
\caption{Comparison results on the ShanghaiTech dataset.}
\label{game_sha}
\begin{tabular}{|c|c|c|c|c|}
  \hline
   & \multicolumn{2}{|c|}{Part-A} & \multicolumn{2}{|c|}{Part-B} \\
  \hline
  Method & MAE & MSE & MAE & MSE \\
  \hline
  \hline
  Crowd-CNN~\cite{zhang2015cross} & 181.8 & 277.7 & 32.0 & 49.8 \\
  \hline
  MCNN~\cite{zhang2016single} & 110.2 & 173.2 & 26.4 &41.3 \\
  \hline
  Cascaded-MTL~\cite{sindagi2017cnn} & 101.3 & 152.4 & 20.0 & 31.1 \\
  \hline
  Switch-CNN~\cite{sam2017switching} & 90.4 & 135.0 & 21.6 &33.4 \\
  \hline
  CP-CNN~\cite{sindagi2017generating} & 73.6 & \textbf{106.4} & 20.1 & 30.1 \\
  \hline
  CMS-CNN-2 (ours) & 74.2 & 127.6 & 15.0 & 25.8 \\
CMS-CNN-3 (ours) & \textbf{73.0} & 128.5 & \textbf{12.0} & \textbf{22.5} \\
  \hline
\end{tabular}
\end{table}

\noindent \textbf{TRANCOS} We also report our results on another dataset for car counting to validate the effectiveness of the proposed method. TRANCOS~\cite{guerrero2015extremely} is a publicly available dataset which contains 1244 images of different traffic scenes obtained by surveillance cameras. An ROI map is also provided for each image. We strictly follow the experimental setup proposed in~\cite{guerrero2015extremely} for training and testing, where there are 403, 420 and 421 images are settled for train, validation and test, respectively. 

Table~\ref{game_trancos} reports the comparison performance on this dataset with four state-of-the-art approaches: density MESA~\cite{lempitsky2010learning}, regression forest~\cite{fiaschi2012learning}, Hydra CNN~\cite{onoro2016towards} and MCNN~\cite{zhang2016single}. The GAME metric with $L = \{0, 1, 2, 3\}$ is utilized for evaluation. Across all the levels of GAME, our method achieves the best results compared to other approaches. There is another work~\cite{zhang2017understanding} reporting their GAME$\sim$0 result of 5.31 on this dataset. However, the other three metrics (GAME$\sim$1, 2, 3) are unavailable for direct and effective comparison. A qualitative result for a sample image from the TRANCOS dataset is shown in Fig.~\ref{fig_subsequent_results} (the third column). It can be seen that the model is able to generate accurate global counting errors with obvious improvements stage-by-stage to become consistency with ground-truth density map.
%
\begin{table}[!htbp]
\centering
\caption{Comparison results of GAME on the TRANCOS dataset.}
\label{game_trancos}
\begin{tabular}{|c|c|c|c|c|c}
  \hline
  Method & GAME 0 & GAME 1 & GAME 2 & GAME 3 \\
\hline
  regression forest~\cite{fiaschi2012learning} & 17.8 &20.1 & 23.6 &26 \\
  \hline
  density MESA~\cite{lempitsky2010learning} & 13.8 & 16.7 & 
  20.7 & 24.4 \\
  \hline
  Hydra CNN~\cite{onoro2016towards} & 11 & 13.7 & 16.7 & 19.3 \\
  \hline
  MCNN~\cite{zhang2016single} & 9.9 & 13 & 15.1 & 17.6 \\
  \hline
CMS-CNN-2 (ours) & 7.79 & 9.81 & 11.57 & 13.69\\
  CMS-CNN-3 (ours) & \textbf{7.2} & \textbf{9.7} & \textbf{11.4} & \textbf{13.5}\\
  \hline
\end{tabular}
\end{table}


\section{Conclusions}
We propose a joint solution to address the local inconsistency problem of existing density map predictions from two aspects. We exploit a different formulation to stack multiple plain CNNs. Benefited from the internal multi-stage inference, the feature map is repeatedly evaluated and thus the density map can be refined to approach the ground-truth density distributions. To further refine the density map, we propose a grid loss function. With local-region-level supervisions, the model is constrained to correct density values which violate the local counts. Extensive experiments on two public counting benchmarks and comparisons with recent state-of-the-art approaches demonstrate the effectiveness of the proposed method. 

%
%
%
\bibliographystyle{splncs04}
\bibliography{mybibliography}

\begin{thebibliography}{10}
\providecommand{\url}[1]{\texttt{#1}}
\providecommand{\urlprefix}{URL }
\providecommand{\doi}[1]{https://doi.org/#1}

\bibitem{chan2008privacy}
Chan, A.B., Liang, Z.S.J., Vasconcelos, N.: Privacy preserving crowd
  monitoring: Counting people without people models or tracking. In: Computer
  Vision and Pattern Recognition, 2008. CVPR 2008. IEEE Conference on.
  pp.~1--7. IEEE (2008)

\bibitem{chen2012feature}
Chen, K., Loy, C.C., Gong, S., Xiang, T.: Feature mining for localised crowd
  counting. In: BMVC. vol.~1, p.~3 (2012)

\bibitem{cirecsan2012multi}
Cire{\c{s}}an, D., Meier, U., Schmidhuber, J.: Multi-column deep neural
  networks for image classification. arXiv preprint arXiv:1202.2745  (2012)

\bibitem{fiaschi2012learning}
Fiaschi, L., K{\"o}the, U., Nair, R., Hamprecht, F.A.: Learning to count with
  regression forest and structured labels. In: Pattern Recognition (ICPR), 2012
  21st International Conference on. pp. 2685--2688. IEEE (2012)

\bibitem{gao2016people}
Gao, C., Li, P., Zhang, Y., Liu, J., Wang, L.: People counting based on head
  detection combining adaboost and cnn in crowded surveillance environment.
  Neurocomputing  \textbf{208},  108--116 (2016)

\bibitem{guerrero2015extremely}
Guerrero-G{\'o}mez-Olmedo, R., Torre-Jim{\'e}nez, B., L{\'o}pez-Sastre, R.,
  Maldonado-Basc{\'o}n, S., O{\~n}oro-Rubio, D.: Extremely overlapping vehicle
  counting. In: Iberian Conference on Pattern Recognition and Image Analysis.
  pp. 423--431. Springer (2015)

\bibitem{he2014spatial}
He, K., Zhang, X., Ren, S., Sun, J.: Spatial pyramid pooling in deep
  convolutional networks for visual recognition. In: European Conference on
  Computer Vision. pp. 346--361. Springer (2014)

\bibitem{kong2006viewpoint}
Kong, D., Gray, D., Tao, H.: A viewpoint invariant approach for crowd counting.
  In: 18th International Conference on Pattern Recognition (ICPR'06). vol.~3,
  pp. 1187--1190. IEEE (2006)

\bibitem{Kumagai2018}
Kumagai, S., Hotta, K., Kurita, T.: Mixture of counting cnns. Machine Vision
  and Applications  (Jul 2018). \doi{10.1007/s00138-018-0955-6},
  \url{https://doi.org/10.1007/s00138-018-0955-6}

\bibitem{lee2015deeply}
Lee, C.Y., Xie, S., Gallagher, P., Zhang, Z., Tu, Z.: Deeply-supervised nets.
  In: Artificial Intelligence and Statistics. pp. 562--570 (2015)

\bibitem{lempitsky2010learning}
Lempitsky, V., Zisserman, A.: Learning to count objects in images. In: Advances
  in Neural Information Processing Systems. pp. 1324--1332 (2010)

\bibitem{li2016iterative}
Li, K., Hariharan, B., Malik, J.: Iterative instance segmentation. In:
  Proceedings of the IEEE Conference on Computer Vision and Pattern
  Recognition. pp. 3659--3667 (2016)

\bibitem{lin2010shape}
Lin, Z., Davis, L.S.: Shape-based human detection and segmentation via
  hierarchical part-template matching. IEEE Transactions on Pattern Analysis
  and Machine Intelligence  \textbf{32}(4),  604--618 (2010)

\bibitem{long2015fully}
Long, J., Shelhamer, E., Darrell, T.: Fully convolutional networks for semantic
  segmentation. In: Proceedings of the IEEE Conference on Computer Vision and
  Pattern Recognition. pp. 3431--3440 (2015)

\bibitem{newell2016stacked}
Newell, A., Yang, K., Deng, J.: Stacked hourglass networks for human pose
  estimation. In: European Conference on Computer Vision. pp. 483--499.
  Springer International Publishing (2016)

\bibitem{onoro2016towards}
Onoro-Rubio, D., L{\'o}pez-Sastre, R.J.: Towards perspective-free object
  counting with deep learning. In: European Conference on Computer Vision. pp.
  615--629. Springer (2016)

\bibitem{pham2015count}
Pham, V.Q., Kozakaya, T., Yamaguchi, O., Okada, R.: Count forest: Co-voting
  uncertain number of targets using random forest for crowd density estimation.
  In: Proceedings of the IEEE International Conference on Computer Vision. pp.
  3253--3261 (2015)

\bibitem{qin2016joint}
Qin, H., Yan, J., Li, X., Hu, X.: Joint training of cascaded cnn for face
  detection. In: Proceedings of the IEEE Conference on Computer Vision and
  Pattern Recognition. pp. 3456--3465 (2016)

\bibitem{sam2017switching}
Sam, D.B., Surya, S., Babu, R.V.: Switching convolutional neural network for
  crowd counting. arXiv preprint arXiv:1708.00199  (2017)

\bibitem{sidla2006pedestrian}
Sidla, O., Lypetskyy, Y., Brandle, N., Seer, S.: Pedestrian detection and
  tracking for counting applications in crowded situations. In: 2006 IEEE
  International Conference on Video and Signal Based Surveillance. pp. 70--70.
  IEEE (2006)

\bibitem{sindagi2017cnn}
Sindagi, V.A., Patel, V.M.: Cnn-based cascaded multi-task learning of
  high-level prior and density estimation for crowd counting. In: Advanced
  Video and Signal Based Surveillance (AVSS), 2017 14th IEEE International
  Conference on. pp.~1--6. IEEE (2017)

\bibitem{sindagi2017generating}
Sindagi, V.A., Patel, V.M.: Generating high-quality crowd density maps using
  contextual pyramid cnns. In: IEEE International Conference on Computer Vision
  (2017)

\bibitem{sindagi2017survey}
Sindagi, V.A., Patel, V.M.: A survey of recent advances in cnn-based single
  image crowd counting and density estimation. Pattern Recognition Letters
  (2017)

\bibitem{vedaldi2015matconvnet}
Vedaldi, A., Lenc, K.: Matconvnet: Convolutional neural networks for matlab.
  In: Proceedings of the 23rd ACM international conference on Multimedia. pp.
  689--692. ACM (2015)

\bibitem{xie2018microscopy}
Xie, W., Noble, J.A., Zisserman, A.: Microscopy cell counting and detection
  with fully convolutional regression networks. Computer methods in
  biomechanics and biomedical engineering: Imaging \& Visualization
  \textbf{6}(3),  283--292 (2018)

\bibitem{zhang2015cross}
Zhang, C., Li, H., Wang, X., Yang, X.: Cross-scene crowd counting via deep
  convolutional neural networks. In: Proceedings of the IEEE Conference on
  Computer Vision and Pattern Recognition. pp. 833--841 (2015)

\bibitem{zhang2017understanding}
Zhang, S., Wu, G., Costeira, J.P., Moura, J.M.: Understanding traffic density
  from large-scale web camera data. arXiv preprint arXiv:1703.05868  (2017)

\bibitem{zhang2016single}
Zhang, Y., Zhou, D., Chen, S., Gao, S., Ma, Y.: Single-image crowd counting via
  multi-column convolutional neural network. In: Proceedings of the IEEE
  Conference on Computer Vision and Pattern Recognition. pp. 589--597 (2016)

\end{thebibliography}

\end{document}